\documentclass[conference]{IEEEtran}
\IEEEoverridecommandlockouts
\usepackage{booktabs}
\usepackage{cite}
\usepackage{amsmath,amssymb,amsfonts}
\usepackage{algorithmic}
\usepackage{graphicx}
\usepackage{textcomp}
\usepackage{xcolor}
\def\BibTeX{{\rm B\kern-.05em{\sc i\kern-.025em b}\kern-.08em
    T\kern-.1667em\lower.7ex\hbox{E}\kern-.125emX}}
\begin{document}

\title{Exploring Advances in Transformers and CNN for Skin Lesion Diagnosis on Small Datasets
}


\author{\IEEEauthorblockN{Leandro M. de Lima}
\IEEEauthorblockA{\textit{Graduate Program in Computer Science, PPGI
} \\
\textit{Federal University of Espirito Santo, UFES}\\
Vitória, Brazil \\
leandro.m.lima@ufes.br}
\and
\IEEEauthorblockN{Renato A. Krohling}
\textit{Graduate Program in Computer Science, PPGI}\\
\IEEEauthorblockA{\textit{Production Engineering Department, LABCIN} \\
\textit{Federal University of Espirito Santo, UFES}\\
Vitória, Brazil \\
rkrohling@inf.ufes.br}
}

\maketitle

\begin{abstract}
Skin cancer is one of the most common types of cancer in the world. Different computer-aided diagnosis systems have been proposed to tackle skin lesion diagnosis, most of them based in deep convolutional neural networks. However, recent advances in computer vision achieved state-of-art results in many tasks, notably Transformer-based networks. We explore and evaluate advances in computer vision architectures, training methods and multimodal feature fusion for skin lesion diagnosis task. Experiments show that PiT ($0.800 \pm 0.006$), CoaT ($0.780 \pm 0.024$) and ViT ($0.771 \pm 0.018$) backbone models with MetaBlock fusion achieved state-of-art results for balanced accuracy metric in PAD-UFES-20 dataset.
\end{abstract}


\section{Introduction}

A third of cancer diagnoses in the world are skin cancer diagnoses according to the World Health Organization (WHO).  In order to diagnose skin cancer, dermatologists screen the skin lesion, assess the patient clinical information, and use their experience to classify the lesion \cite{pacheco2020impact}. The high incidence rate and the lack of experts and medical devices, specifically in rural areas \cite{feng2018comparison}  and emerging countries \cite{scheffler2008forecasting}, have increased the demand for computer-aided diagnosis (CAD) systems for skin cancer. 

Over the past decades, different computer-aided diagnosis (CAD) systems have been proposed to tackle skin cancer detection \cite{takiddin2021artificial, das2021artificial}. In general, these systems are based on clinical information from the patient and information extracted from lesion images \cite{Zhou2021FusionMAT}. Image features and features extracted from clinical information need to be merged. This task is known as multimodal data fusion.

The use of neural network has become the de-facto standard as a backbone for extracting visual features in various tasks. Deep Convolutional Neural Network (CNN) based architectures are widely used for this. However, recently, Transformer-based networks stand out for achieving comparable performance in various tasks. Other recent advances, such as new training methods \cite{sirotkin2021improved} and the proposal of new architectures \cite{karthik2022eff2net}, may also contribute to an improvement in the performance of the skin lesion diagnosis task.

This work has as main objectives investigate the performance of the most recent architectures for computer vision in the problem of skin lesion detection. We also investigate the performance of feature fusion methods for the problem of skin lesion detection. Additionally, we investigate how these new architectures and these fusion methods can be integrated and the resulting performance is compared.

The following are the key research contributions of the proposed work.

\begin{itemize}
\item we conduct extensive experiments on the open dataset PAD-UFES-20\cite{pacheco2020pad} and achieved performance comparable to state-of-art methods.

\item we show that Transformer-based image feature extractors achieve competitive performance against CNN-based backbones to skin lesion diagnosis task.

\item we show that Transformer-based image extracted features can be fused with clinical information using already existent fusion methods.

\item we show that recent training methods (distillation and semi-weakly supervised training) and recent architectures (ResNeXt, PiT and CoaT) effectively can improve performance in skin lesion diagnosis task.
\end{itemize}

\section{Literature Review}

\subsection{CNN-based vision backbones}

A Deep Convolutional Neural Network (CNN) model consists of several convolution layers followed by activation functions and pooling layers. Additionally, it has several fully connected layers before prediction. It comes into deep structure to facilitate filtering mechanisms by performing convolutions in multi-scale feature maps, leading to highly abstract and discriminative features \cite{feng2019computer}. 

Several architectures have been developed since the AlexNet architecture, considered as the foundation work of modern deep CNN, with great emphasis on architectures such as ResNet, DenseNet and EfficientNet \cite{bhatt2021cnn}. More recently, improvements in these networks have been proposed as in ResNet V2 \cite{he2016identity}, ResNexT \cite{xie2017aggregated}  and EfficientNet V2 \cite{tan2021efficientnetv2}. In addition to the development of architectures, there are proposals for new mechanisms (e.g. Efficient Channel Attention\cite{wang2020eca}) and new training methods (e.g. Semi-weakly Supervised Learning \cite{yalniz2019billion}, Distillation \cite{hinton2015distilling}).

\subsection{Transformer-based vision backbones}

\begin{figure}[htbp]
\centerline{\includegraphics[scale=0.27]{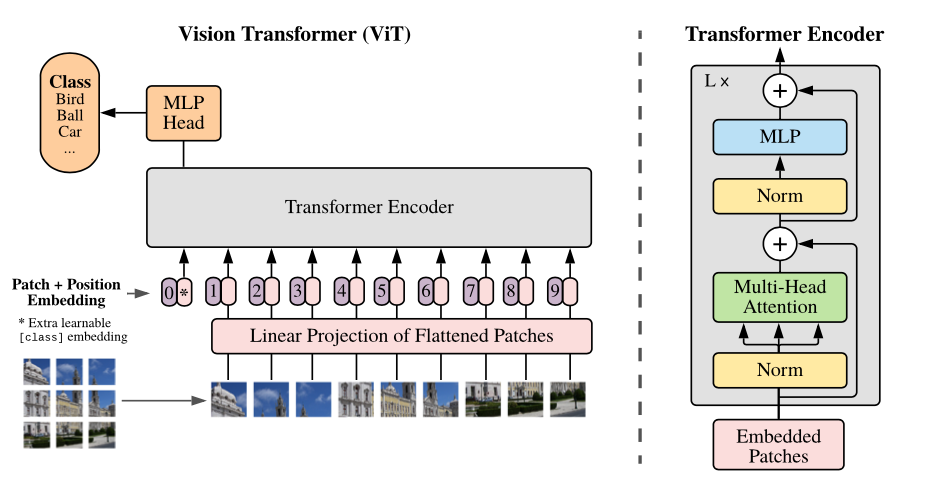}}
\caption{ViT architecture overview (left) and details of Transformer Encoder (right). Image from \cite{dosovitskiy2020image}.}
\label{fig:vit}
\end{figure}

With the great success of Transformer-based architectures for NLP tasks, there has recently been a great interest in researching Transformer-based architectures for computing vision \cite{khan2021transformers, han2020survey, liu2021survey, xu2022transformers}. Vision Transformer (ViT) \cite{dosovitskiy2020image} was one of the pioneers to present comparable results with the CNN architectures, until then dominant. Its architecture is heavily based on the original Transformer model \cite{vaswani2017attention} and it process the input image as a sequence of patches, as shown in Figure \ref{fig:vit}.

Inspired by the advances achieved by the ViT model, various models and modifications (e.g. TNT \cite{han2021transformer}, Swin \cite{liu2021swin} / SwinV2 \cite{liu2021swinv2}, CrossViT \cite{chen2021crossvit}, XCiT \cite{el2021xcit}, PiT \cite{heo2021rethinking}, CaiT \cite{touvron2021going}
) were proposed and presented promising results. In addition, several improvements in training methods (e.g. BeiT \cite{bao2021beit}, DeiT \cite{touvron2021deit}, iBOT \cite{zhou2021ibot}, DINO \cite{caron2021dino}) have also contributed to an improvement in the performance of Transformer-based models.

An interesting ability of Transformer-based models is that ViTs trained on ImageNet exhibit higher shape-bias in comparison to similar capacity CNN models \cite{naseer2021intriguing}. Transformer-based models can reach human-level shape-bias performance when trained on a stylized version of ImageNet (SIN \cite{geirhos2018imagenet}) and they can even model contradictory cues (as in shape/texture-bias) with distinct tokens \cite{naseer2021intriguing}.

\subsection{Multimodal Fusion}

There are some aggregation-based multimodal fusion approaches. The most common method of multimodal fusion is an aggregation via concatenation of features\cite{pacheco2020impact}. This fusion method consists of concatenating the features into a single tensor with all features. 

Channel-Exchanging-Network (CEN)\cite{Wang2020FusionCEN}, a parameter-free multimodal fusion framework, is another approach that proposes an exchanging in CNN channels of sub-networks. There is an intrinsic limitation in it as the framework assumes that all sub-networks have a CNN architecture.

An interesting alternative is the Metadata Processing Block
(MetaBlock) \cite{Pacheco2021FusionMetaBlock}, that is an attention-based method that uses a LSTM-like gates to enhance the metadata into the feature maps extracted from an image for skin cancer classification. This method main limitation is that it is proposed for only one feature map and one metadata source and there is no clear information how to scale it for multiple sources.

Also, there is MetaNet \cite{Li2020FusionMetaNet}, a multiplication-based data fusion to make the metadata directly interact with the visual features. It proposes use the metadata to control the importance of each feature channel at the last convolutional layer.

Mutual Attention Transformer (MAT) \cite{Zhou2021FusionMAT} uses an attention-based multimodal fusion method (Transformer and Fusion unit) that is inspired in transformer architecture. Being proposed for only two features sources is a limitation in it. MAT presents a guided-attention module which has some similarities with the cross-attention module in the Cross-Attention Fusion \cite{chen2021crossvit}. Cross-Attention Fusion is the proposed method for multi-scale feature fusion in CrossViT architecture.

\section{Proposed Evaluation Methodology}

Focused on investigating performance, techniques and recent architectures that bring advantage to the skin lesion classification task, we selected 20 models with $30$M parameters or less and that needed less than $12$GB GPU memory (due hardware limitation). In case of identical architectures and techniques, we use only the best model of them. We took the $20$ best top-$1$ validation scores in TIMM pre-trained models collection\cite{rw2019timm} evaluated in ImageNet ''Reassessed Labels'' (ImageNet-ReaL)\cite{beyer2020imagenetreal}, the usual ImageNet-1k validation set with a fresh new set of labels intended to improve on mistakes in the original annotation process. All selected models are listed in Table \ref{table:top20_param} with their number of parameters and what kind of architecture they are based on.

\begin{table}[htbp]
\caption{Number of parameters and base architecture of the top-$20$ selected models}
\centering
\begin{tabular}{l|c|c|c} 
\toprule
\multicolumn{1}{c|}{\textbf{Model}} & \textbf{Based} & \textbf{Param.} & \textbf{Ref.}  \\ 
\hline
cait\_xxs24\_384                    & Transformer

 & {12.03 M} & \cite{touvron2021going}                    \\
gc\_efficientnetv2\_rw\_t           & CNN
& {13.68 M} & \cite{tan2021efficientnetv2, cao2019gcnet}                    \\
rexnet\_200                         & CNN
& {16.37 M} & \cite{han2021rethinking}                    \\
tf\_efficientnet\_b4\_ns            & CNN
& {19.34 M} & \cite{tan2019efficientnet, xie2020self}                    \\
regnety\_032                        & CNN
& {19.44 M} & \cite{radosavovic2020designing}                     \\
coat\_lite\_small                   & Transformer

& {19.84 M} & \cite{xu2021co}                    \\
tf\_efficientnetv2\_s\_in21ft1k     & CNN
& {21.46 M} & \cite{tan2021efficientnetv2}                    \\
vit\_small\_patch16\_384            & Transformer

& {22.20 M} & \cite{dosovitskiy2020image, steiner2021train}                    \\
halo2botnet50ts\_256                & Hybrid

& {22.64 M} & \cite{srinivas2021bottleneck, vaswani2021scaling}                    \\
halonet50ts                         & Hybrid

& {22.73 M} & \cite{vaswani2021scaling, he2016deep}                    \\
tnt\_s\_patch16\_224                & Transformer
& {23.76 M} & \cite{han2021transformer}                    \\
pit\_s\_distilled\_224              & Transformer
& {24.04 M} & \cite{heo2021rethinking}                    \\
twins\_svt\_small                   & Transformer
& {24.06 M} & \cite{chu2021twins}                    \\
eca\_nfnet\_l0                      & CNN
& {24.14 M} & \cite{brock2021high, wang2020eca}                    \\
swsl\_resnext50\_32x4d              & CNN
& {25.03 M} & \cite{xie2017aggregated, yalniz2019billion}                    \\
resnetv2\_50x1\_bit\_distilled      & CNN
& {25.55 M} & \cite{he2016identity, beyer2021knowledge, kolesnikov2020big}                    \\
ecaresnet50t                        & CNN
& {25.57 M} & \cite{he2016deep}                    \\
xcit\_small\_12\_p16\_384\_dist     & Transformer
& {26.25 M} & \cite{el2021xcit}                    \\
regnetz\_d8                         & CNN
& {27.58 M} & \cite{dollar2021fast}                    \\
crossvit\_15\_dagger\_408           & Transformer
& {28.50 M} & \cite{chen2021crossvit}                    \\
\bottomrule
\end{tabular}
\label{table:top20_param}
\end{table}

\begin{figure}[htbp]
\centerline{\includegraphics[scale=0.5]{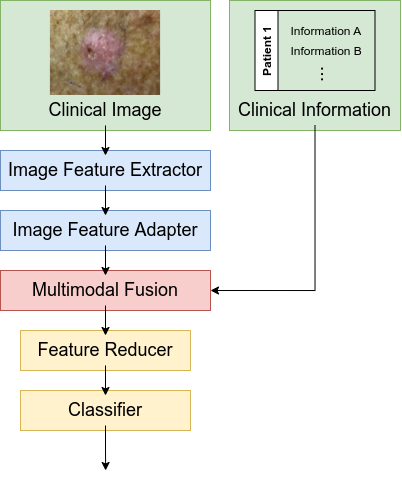}}
\caption{Fusion architecture based on \cite{Pacheco2021FusionMetaBlock}.}
\label{fig:method_arch}
\end{figure}

For the evaluation of all models, we based our methodology on the one proposed by \cite{pacheco2020impact}, as shown in Figure \ref{fig:method_arch}. The input data is composed of a clinical image of the lesion and clinical information. The clinical image features are extracted by a backbone model pre-trained for general image tasks. Since we test a large range of different architectures, we need a image feature adapter to reformat it to a standard shape. That adapter consist of a 2D adaptive average pooling layer for hybrid or CNN-based image features. For transformers-based image features, the adapter only selects the class token and discard other image information, except for CrossViT that big and small class tokens are concatenated and selected. The adapter also has a flatten layer after it to properly reformat the image features to a standard shape. Next, a fusion block add clinical information to reformatted image features. In this paper we analyze four alternatives to the fusion block, i.e., Concatenation, MetaBlock, MetaNet, MAT (Transformer and Fusion unit only). For a fair comparison, before the classifier, we apply a feature reducer to keep the classifier input the same size to all models, as most of those image feature extractor models have different image features size.

The selected pre-trained models are fine-tuned with supervised learning on PAD-UFES-20 dataset \cite{pacheco2020pad}. The dataset has $2298$ samples of $6$ types of skin lesions, consisting of a clinical image collected from smartphone devices and a set of patient clinical data containing $21$ features. The skin lesions are Basal Cell Carcinoma, Squamous Cell Carcinoma, Actinic Keratosis, Seborrheic Keratosis, Melanoma, and Nevus. The clinical features include patient’s age, skin lesion location, Fitzpatrick skin type, skin lesion diameter, family background, cancer history, among others. That dataset was chosen precisely because it deals with multimodal data and is one of the few containing this type of information for skin lesion diagnosis. Most deal only with dermoscopic images without clinical information of the lesion.

\section{Experiments}
Our experiments follows the setup used in \cite{Pacheco2021FusionMetaBlock}. Training runs for $150$ epochs using batch size $30$ and reducer block size of $90$. An early stop strategy is set if training for $15$ epochs without improvement. We use the SGD optimizer (initial learning rate is $0.001$, momentum is $0.9$ and weight decay is $0.001$) with reduce learning rate on plateau strategy (patience is $10$, reduce factor is $0.1$ and lower bound on the learning rate is $10^{-6}$). For evaluation, we adopted $5$-fold cross-validation. 

\subsection{General Analysis}
Table \ref{table:top20_result} list the balanced accuracy (BCC) and area under the ROC curve (AUC) performance of each pre-trained backbone model using Concatenation fusion. The model ''resnetv2\_50x1\_bit\_distilled'' achieved the best balanced accuracy and area under the ROC curve performance with $0.765 \pm 0.013$ and $0.934 \pm 0.002$. The ''tf\_efficientnetv2\_s\_in21ft1k'' backbone model achieved the same best area under the ROC curve performance $0.934 \pm 0.004$. 

\begin{table}
\centering
\caption{Top-20 selected models concatenation fusion performance ordered by balanced accuracy (BCC). Mean and standard deviation of BCC and area under the ROC curve (AUC) metrics.}
\label{table:top20_result}
\begin{tabular}{l|ll} 
\toprule
\multicolumn{1}{c|}{\textbf{Model}} & \multicolumn{1}{c}{\textbf{BCC}}   & \multicolumn{1}{c}{\textbf{AUC}}    \\ 
\hline
resnetv2\_50x1\_bit\_distilled      & $\textbf{0.765} \pm 0.013$ & $\textbf{0.934} \pm 0.002$  \\
pit\_s\_distilled\_224              & $0.763 \pm 0.025$ & $0.928 \pm 0.009$  \\
coat\_lite\_small                   & $0.759 \pm 0.024$ & $0.929 \pm 0.002$  \\
vit\_small\_patch16\_384            & $0.751 \pm 0.017$ & $0.926 \pm 0.011$  \\
regnety\_032                        & $0.748 \pm 0.018$ & $0.927 \pm 0.010$  \\ 
\hline
twins\_svt\_small                   & $0.747 \pm 0.027$ & $0.927 \pm 0.005$  \\
ecaresnet50t                        & $0.742 \pm 0.039$ & $0.924 \pm 0.008$  \\
tf\_efficientnetv2\_s\_in21ft1k     & $0.741 \pm 0.027$ & $\textbf{0.934} \pm 0.004$  \\
regnetz\_d8                         & $0.739 \pm 0.021$ & $0.930 \pm 0.007$  \\
gc\_efficientnetv2\_rw\_t           & $0.739 \pm 0.026$ & $0.926 \pm 0.010$  \\
eca\_nfnet\_l0                      & $0.736 \pm 0.037$ & $0.926 \pm 0.007$  \\
swsl\_resnext50\_32x4d              & $0.731 \pm 0.028$ & $0.925 \pm 0.004$  \\
rexnet\_200                         & $0.728 \pm 0.028$ & $0.928 \pm 0.007$  \\
xcit\_small\_12\_p16\_384\_dist     & $0.727 \pm 0.032$ & $0.921 \pm 0.010$  \\
tf\_efficientnet\_b4\_ns            & $0.726 \pm 0.017$ & $0.923 \pm 0.016$  \\
tnt\_s\_patch16\_224                & $0.725 \pm 0.025$ & $0.925 \pm 0.007$  \\
crossvit\_15\_dagger\_408           & $0.718 \pm 0.033$ & $0.919 \pm 0.008$  \\
halo2botnet50ts\_256                & $0.701 \pm 0.023$ & $0.916 \pm 0.011$  \\
cait\_xxs24\_384                    & $0.660 \pm 0.027$ & $0.910 \pm 0.006$  \\
halonet50ts                         & $0.644 \pm 0.054$ & $0.903 \pm 0.015$  \\
\bottomrule
\end{tabular}
\end{table}

From the results listed in Table \ref{table:top20_result} we concluded that the use of Transformer-based backbone can help improve skin lesion diagnosis performance, since in the top 5 balanced accuracy performance three of them are Transformer-based architectures and two of them are CNN-based architectures. It can also be noted the presence of two architectures that used the model distillation \cite{hinton2015distilling} technique for training. 

Next, we will detail the main architectures and present the main characteristics that may have contributed to its good result in the top-$5$ pre-trained models.

\subsubsection{Model Distillation}
Also known as Knowledge Distillation\cite{hinton2015distilling}, refers to the training paradigm in which a student model leverages ''soft'' labels coming from a strong teacher network. The output vector of the teacher’s softmax function is used rather than just the maximum of scores, which gives a “hard” label. This process can be seen as a way of compressing the teacher model into a reduced student model. In Transformer-based models, a distillation token can be added to the model, along class and patch tokens, and used to improve learning\cite{touvron2021deit}.

\subsubsection{ResNet V2}

The Residual Neural Network (ResNet) V2 \cite{he2016identity} mainly focuses on making the second non-linearity as an identity mapping by removing the last ReLU activation function, after the addition layer, in the residual block. That is, using the pre-activation of weight layers instead of post-activation.

\begin{figure}[htbp]
\centerline{\includegraphics[scale=0.45]{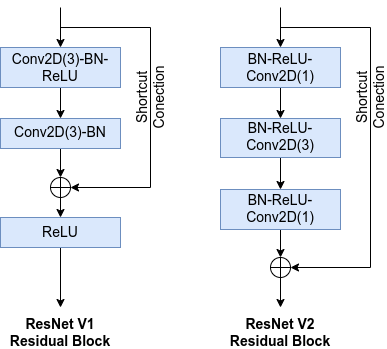}}
\caption{Differences in residual block of ResNet versions. Based on image from \cite{he2016identity}.}
\label{fig:resnet_version}
\end{figure}

The arrangement of the layers in the residual block moves the batch normalization and ReLU activation to comes before 2D convolution, as shown in Figure \ref{fig:resnet_version}.

\subsubsection{PiT}

\begin{figure}[htbp]
\centerline{\includegraphics[scale=0.25]{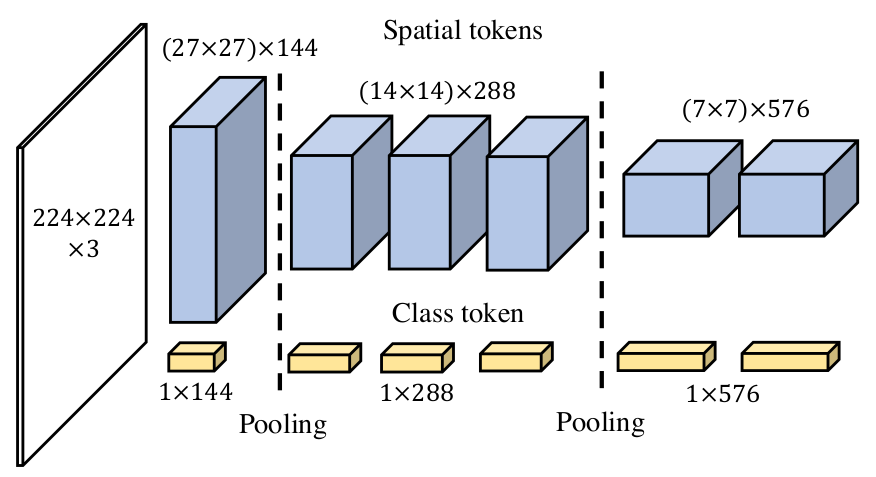}}
\caption{Pooling-based Vision Transformer (PiT)  architecture. Image from \cite{heo2021rethinking}.}
\label{fig:pit}
\end{figure}

Pooling-based Vision Transformer (PiT)  architecture \cite{heo2021rethinking} is inspired in ResNet-style dimensions settings to improve the model performance, as shown in Figure \ref{fig:pit}. PiT uses a newly designed pooling layer based on depth-wise convolution to achieve channel multiplication and spatial reduction. 

\subsubsection{CoaT}

Co-scale conv-attentional image Transformers (CoaT) \cite{xu2021co} is a Transformer-based model equipped with co-scale and conv-attentional mechanisms. It empowers image Transformers with
enriched multi-scale and contextual modeling capabilities. The conv-attentional module realize relative position embeddings with convolutions in the factorized attention module, which improves computational efficiency.

\begin{figure}[htbp]
\centerline{\includegraphics[scale=0.55]{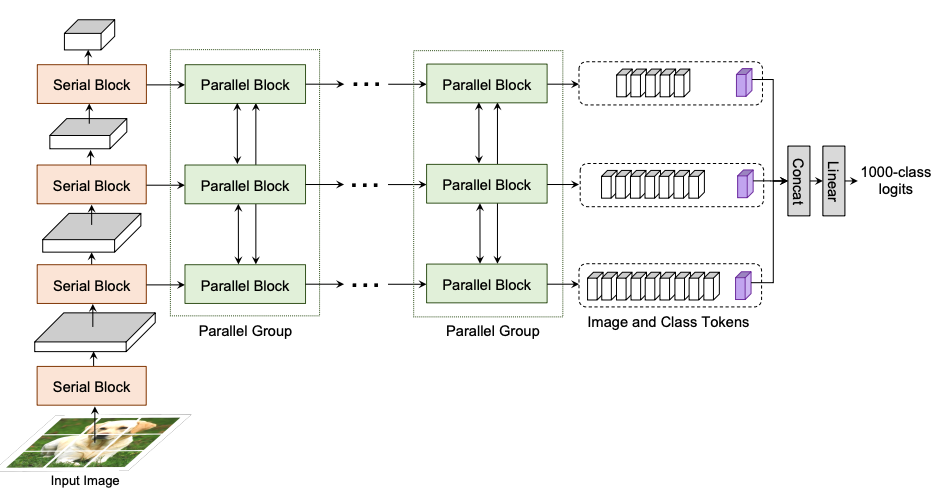}}
\caption{Co-scale conv-attentional image Transformers (CoaT) architecture. Image from \cite{xu2021co}.}
\label{fig:coat}
\end{figure}

CoaT designs a series of serial and parallel blocks to realize the co-scale mechanism as shown in Figure \ref{fig:coat}. The serial block models image representations in a reduced resolution. Then, CoaT realizes the co-scale mechanism between parallel blocks in each parallel group.

\subsubsection{ViT}
Due the success of Transformer-based architectures for NLP tasks, there has recently been a great interest in researching Transformer-based architectures for computing vision \cite{khan2021transformers, han2020survey, liu2021survey, xu2022transformers}. Vision Transformer (ViT) \cite{dosovitskiy2020image} was one of the pioneers to present comparable results with the CNN architectures, until then dominant. Its architecture is heavily based on the original Transformer model \cite{vaswani2017attention} and it process the input image as a sequence of patches, as shown in Figure \ref{fig:vit}.

\subsubsection{RegNetY}

\begin{figure}[htbp]
\centerline{\includegraphics[scale=0.33]{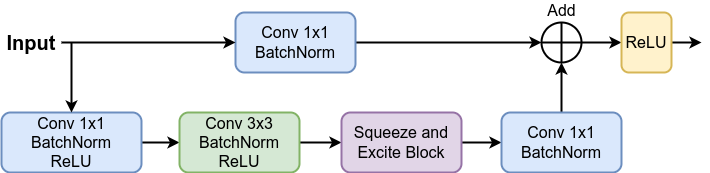}}
\caption{RegNetY Y block diagram (when $stride > 1$).}
\label{fig:yblock}
\end{figure}

RegNetY \cite{radosavovic2020designing} is one of a family of models proposed by a methodology to design network design spaces, where a design space is a parametrized set of possible model architectures. Each RegNet network consists of a stem, followed by the network body that performs the bulk of the computation, and then a head (average pooling followed by a fully connected layer) that predicts $n$ output classes. The network body is composed of a sequence of stages that operate at progressively reduced resolution. Each stage consists of a sequence of identical blocks, except the first block which uses stride-two convonvolution. While the general structure is simple, the total number of possible network configurations is vast. For RegNetY models, the blocks in each stage are based on the standard residual bottleneck block with group convolution and with the addition of a Squeeze and Excite Block after the group convolution, as shown in Figure \ref{fig:yblock}.

\subsection{Feature Fusion Analysis}

Aiming to analyze feature extraction models alternatives and their impact for the fusion of clinical image and clinical information features, the top-$5$ models with the best balanced accuracy performance in Table \ref{table:top20_result} is compared using multiple fusion methods. Each selected model is evaluated using the fusion methods: Concatenation (already presented in Table \ref{table:top20_result}), MetaBlock, MetaNet and MAT (Transformer and Fusion unit only). As baseline we use the best performance result in \cite{pacheco2020impact}, which is a balanced accuracy of $0.770 \pm 0.016$ (EfficientNet-B4 with MetaBlock fusion) and area under the ROC curve (AUC) of $0.945 \pm 0.005$ (EfficientNet-B4 with Concatenation fusion). Currently, those are the best results published for PAD-UFES-20 dataset.

Results listed in Table \ref{table:top5_fusion} show that most of CNN-based models evaluated have a better BCC and AUC with Concatenation Fusion and most of Transformer-based models evaluated have a better BCC and AUC with MataBlock fusion. Analysing balanced accuracy metric, ''pit\_s\_distilled\_224'' ($0.800 \pm 0.006$), ''coat\_lite\_small'' ($0.780 \pm 0.024$) and ''vit\_small\_patch16\_384'' ($0.771 \pm 0.018$) models with MetaBlock fusion achieved state-of-art results. For area under the ROC curve metric, ''pit\_s\_distilled\_224'' ($0.941 \pm 0.006$) model with MetaBlock fusion has the best mean result among proposed models, but EfficientNet-B4 with Concatenation fusion baseline has the best AUC among all of them. All models highest average AUC (in bold at Table \ref{table:top5_fusion}) do note achieved state-of-art results.

\begin{table}
\centering
\caption{Top-5 selected models with performance for fusion method comparison. In bold is highlighted the highest average for each backbone model.}
\label{table:top5_fusion}
\begin{tabular}{l|l|l} 
\toprule
\multicolumn{1}{c|}{\textbf{Model}} & \multicolumn{1}{c|}{\textbf{BCC}}  & \multicolumn{1}{c}{\textbf{AUC}}    \\ 
\hline
\multicolumn{3}{c}{Baseline \cite{pacheco2020impact}}                                                                                   \\ 
\hline
EfficientNet-B4 w/ MetaBlock        & $\textbf{0.770} \pm 0.016$                  & $0.944 \pm 0.004$                   \\
EfficientNet-B4 w/ Concatenation    & $0.758 \pm 0.012$                  & $\textbf{0.945} \pm 0.005$                   \\ 
\hline
\multicolumn{3}{c}{Concatenation Fusion}                                                                       \\ 
\hline
coat\_lite\_small                   & $0.759 \pm 0.024$ & $0.929 \pm 0.002$  \\
pit\_s\_distilled\_224              & $0.763 \pm 0.025$ & $0.928 \pm 0.009$  \\
regnety\_032                        & $\textbf{0.748} \pm 0.018$ & $\textbf{0.927} \pm 0.010$  \\
resnetv2\_50x1\_bit\_distilled      & $\textbf{0.765} \pm 0.013$ & $0.934 \pm 0.002$  \\
vit\_small\_patch16\_384            & $0.751 \pm 0.017$ & $0.926 \pm 0.011$  \\ 
\hline
\multicolumn{3}{c}{MAT Fusion}                                                                                 \\ 
\hline
coat\_lite\_small                   & $0.685 \pm 0.015$ & $0.909 \pm 0.009$  \\
pit\_s\_distilled\_224              & $0.704 \pm 0.027$ & $0.913 \pm 0.013$  \\
regnety\_032                        & $0.678 \pm 0.012$ & $0.913 \pm 0.006$  \\
resnetv2\_50x1\_bit\_distilled      & $0.716 \pm 0.022$ & $0.917 \pm 0.007$  \\
vit\_small\_patch16\_384            & $0.729 \pm 0.032$ & $0.920 \pm 0.015$  \\ 
\hline
\multicolumn{3}{c}{MetaBlock Fusion}                                                                           \\ 
\hline
coat\_lite\_small                   & $\textbf{0.780} \pm 0.024$ & $\textbf{0.940} \pm 0.002$  \\
pit\_s\_distilled\_224              & $\textbf{0.800} \pm 0.006$ & $\textbf{0.941} \pm 0.006$  \\
regnety\_032                        & $0.705 \pm 0.015$ & $0.920 \pm 0.005$  \\
resnetv2\_50x1\_bit\_distilled      & $0.702 \pm 0.015$ & $0.918 \pm 0.004$  \\
vit\_small\_patch16\_384            & $\textbf{0.771} \pm 0.018$ & $0.936 \pm 0.008$  \\ 
\hline
\multicolumn{3}{c}{MetaNet Fusion}                                                                             \\ 
\hline
coat\_lite\_small                   & $0.732 \pm 0.038$ & $0.927 \pm 0.011$  \\
pit\_s\_distilled\_224              & $0.754 \pm 0.028$ & $0.932 \pm 0.008$  \\
regnety\_032                        & $0.717 \pm 0.013$ & $0.923 \pm 0.010$  \\
resnetv2\_50x1\_bit\_distilled      & $0.752 \pm 0.020$ & $\textbf{0.936} \pm 0.006$  \\
vit\_small\_patch16\_384            & $0.767 \pm 0.021$ & $\textbf{0.938} \pm 0.011$  \\
\bottomrule
\end{tabular}
\end{table}

\section{Conclusion}

Computer-aided diagnosis (CAD) systems for skin cancer has an increasing demand. Computer vision has largely helped the development of efficient predictive models for skin lesion diagnosis. Recent advances in computer vision, as newly architectures and training methods, has provided performance improvements in many tasks. This work seeks to clarify and guide researchers about which architectures and other advances present an effective improvement in skin lesion diagnosis task. It also attempts to elucidate which are the most suitable pre-trained models available for the task. 

The average AUC of ''pit\_s\_distilled\_224'' ($0.941 \pm 0.006$) and ''coat\_lite\_small'' ($0.940 \pm 0.002$) models with MetaBlock fusion achieved competitive results. Experiments also show that
''pit\_s\_distilled\_224'' ($0.800 \pm 0.006$), ''coat\_lite\_small'' ($0.780 \pm 0.024$) and ''vit\_small\_patch16\_384'' ($0.771 \pm 0.018$) models with MetaBlock fusion achieved balanced accuracy metric state-of-art results. It shows that distillation and transformer architectures (PiT, CoaT and ViT) can improve performance in skin lesion diagnosis task. Our findings are not in accordance with that provided by \cite{zhao2021comparison}, as they conclude that CNN-based models obtained better results in small datasets ($840$ images and $21$ labels, smaller dataset than PAD-UFES-20) and here experiments indicate that Transformer-based backbones achieved state-of-art results. 

For future work, one can investigate a fusion method that takes advantage of the specifics of the format of the features extracted by Transformer-based backbones. Novel architectures or training methods can be further investigated and developed. There is also room for improvement in the understand of the impact of the advances in model interpretability and explainability in skin lesion diagnosis.

\bibliographystyle{IEEEtran}
\bibliography{IEEEabrv,bibliography}

\end{document}